\documentclass[10.5pt]{article}

\usepackage{wrapfig}

\usepackage{mathpazo}
\usepackage[utf8]{inputenc}
\usepackage{verbatim}
\usepackage{calc}
\usepackage{mathrsfs}
\usepackage{mathtools}
\usepackage{url}
\usepackage{algorithm2e}
\usepackage{amsmath}
\usepackage{amssymb}
\usepackage{graphicx}
\usepackage{hyperref}
\usepackage{xcolor}
\usepackage{comment}
\usepackage{soul}
\usepackage{booktabs}

\definecolor{forestgreen}{rgb}{0.13, 0.55, 0.13}

\usepackage{tabularx}

\makeatletter

\newcommand{\lyxmathsym}[1]{\ifmmode\begingroup\def\b@ld{bold}
  \text{\ifx\math@version\b@ld\bfseries\fi#1}\endgroup\else#1\fi}

\usepackage{graphics}\usepackage{epsfig}

\usepackage{soul}

\usepackage{graphicx}
\usepackage{booktabs} 

\usepackage{here}

\usepackage{amsmath,amssymb,amsfonts,amsbsy,amsfonts,latexsym}
\usepackage{multirow}
\usepackage{makecell}
\usepackage[labelfont=bf,belowskip=0pt,aboveskip=5pt,tableposition=top]{caption}
\usepackage{xcolor}
\usepackage{colortbl}

\definecolor{colorA}{RGB}{189,201,225}
\definecolor{colorB}{RGB}{103,169,207}
\definecolor{colorC}{RGB}{ 28,144,153}
\definecolor{colorD}{RGB}{  1,108, 89}

\newcolumntype{R}{>{\columncolor{gray!40}}r}
\newcolumntype{L}{>{\columncolor{gray!40}}l}
\newcolumntype{C}{>{\columncolor{gray!40}}c}

\newcommand\blfootnote[1]{%
  \begingroup
  \renewcommand\thefootnote{}\footnote{#1}%
  \addtocounter{footnote}{-1}%
  \endgroup
}

\usepackage{tabularx,colortbl,xcolor}
\usepackage{multirow}
\usepackage[normalem]{ulem}
\useunder{\uline}{\ul}{}

\usepackage{enumitem}

\usepackage{xparse}

\DeclareGraphicsExtensions{.pdf,.png}

\SetKwInput{KwInput}{Input}

\usepackage{longtable}
\usepackage{pgfplots}
\usepackage{outlines}

\usepackage[many]{tcolorbox}
\usepackage{caption}
\usepackage{subcaption}
\usepackage{graphbox} 

\tcbset{
    sharp corners,
    colback = white,
    before skip = 0.2cm,    
    after skip = 0.5cm      
}                           

\definecolor{main}{HTML}{4472C4}    
\definecolor{sub}{HTML}{EBF4FF}     

\newtcolorbox{boxA}{
    enhanced, breakable,
    boxrule = 0pt,
    colback = sub,
    borderline west = {2pt}{0pt}{main}, 
    borderline east = {2pt}{0pt}{main}, 

}

\usepackage{times}
\usepackage{textcomp}

\newcommand{\OURS}{{Tool2Vec}\xspace}
\newcommand{\RERANKER}{{ToolRefiner}\xspace}
\newcommand{\SIDFORMER}{{MLC}\xspace}
\newcommand{\NUMPY}{{NumpyBank}\xspace}
\newcommand{\PANDAS}{{PandasBank}\xspace}
\newcommand{\AWS}{{AWSBank}\xspace}
\newcommand{\OURDATA}{{ToolBank}\xspace}

\newcommand{\highlightrowb}{\rowcolor{teal!15}}
\newcommand{\highlightrow}{\rowcolor{teal!7}}

\newcommand{\graycell}[1]{\cellcolor{teal!7}#1}

\title{Efficient and Scalable Estimation of Tool Representations \\in Vector Space}

\author{
Suhong Moon\blfootnote{Equal contribution}$^{*1}$ \enspace\enspace Siddharth Jha$^{*1}$ \enspace\enspace Lutfi Eren Erdogan$^{1}$ \enspace\enspace Sehoon Kim$^{1}$\\
Woosang Lim$^{2}$\enspace\enspace Kurt Keutzer$^{1}$\enspace\enspace Amir Gholami$^{1,3}$\vspace{3mm}
\\
{$^{1}$~UC Berkeley\qquad $^{2}$~POSCO HOLDINGS\qquad $^{3}$~LBNL}\vspace{3mm}\\
}
\date{}  
\makeatother

\begin{document}

\maketitle

\begin{abstract}
Recent advancements in function calling and tool use have significantly enhanced the capabilities of large language models (LLMs) by enabling them to interact with external information sources and execute complex tasks. 
However, the limited context window of LLMs presents challenges when a large number of tools are available, necessitating efficient methods to manage prompt length while maintaining accuracy. 
Existing approaches, such as fine-tuning LLMs or leveraging their reasoning capabilities, either require frequent retraining or incur significant latency overhead. 
A more efficient solution involves training small models to retrieve the most relevant tools for a given query. 
However, previous methods rely on tool descriptions, leading to suboptimal performance. 
To address this, we propose approaches based on a two-stage retrieval technique. 
In the first stage, candidate tools are retrieved using a fast retriever, incorporating two novel methods: (1) \OURS, which generates usage-driven tool embeddings, and (2) \SIDFORMER, which frames tool retrieval as a multi-label classification problem. 
In the second stage, we introduce \RERANKER, which refines the candidate tools retrieved in the first stage, further enhancing retrieval performance.
While this approach requires domain-specific tool retrieval data, we demonstrate that LLMs can generate high-quality datasets. 
To show this, we create \OURDATA, showcasing that LLMs can generate effective tool retrieval data across various domains.
With our methods, we achieve improvements of up to 27.28 in Recall@K on the ToolBench dataset and 30.5 in Recall@K on \OURDATA. 
The dataset and code are publicly available.\footnote{Code: \url{https://github.com/SqueezeAILab/Tool2Vec}}\footnote{ToolBank Dataset: \url{https://huggingface.co/datasets/squeeze-ai-lab/ToolBank}}
\end{abstract}
\section{Introduction}
Recently, function calling and tool use has emerged as a powerful paradigm for using large language models (LLMs)~\cite{patil2023gorilla,schick2024toolformer,openai2023functioncalling,cai2023large}.
Rather than relying solely on the model's parametric knowledge, function calling and tool use enable the model to interact with the world ~\cite{tiny-agent,chen2024octopus,yao2022react,wang2023voyager,kim2023llmcompiler}. 
This approach allows the model to perform specific tasks, such as accessing information beyond the LLM's knowledge cut-off date, solving complex math problems, and executing complex planning~\cite{trinh2024solving,silver2024generalized,karpas2022mrkl,chen2022codet}.

However, since function calling requires passing in the tool's description and signature into the model's context window, 
it is often infeasible to put information about potentially thousands of functions due to context window limitations.
Additionally, even when using models with longer context windows, long context inference leads to latency, cost overheads and accuracy challenges, necessitating the need for smaller prompts~\cite{kim2023full, jha2024characterizing,tiny-agent,kim2023llmcompiler}.
Therefore, selectively retrieving tools to present to the model can greatly reduce prompt lengths while preserving accuracy.

Several methods have been proposed to address the issue of the limited context window in LLMs when the number of available tools exceeds the model's capacity.
One popular approach is to leverage the reasoning capabilities of LLMs to pre-select the appropriate tools from a large pool~\cite{du2024anytool,shinn2024reflexion,wang2023voyager,yuan2024easytool}.
Despite the LLMs' ability to learn and choose tools effectively, this method incurs significant latency overhead, making it less practical in various use cases where real-time responses are critical.
An alternative and more efficient solution is dense retrieval of tools~\cite{qin2024toolllm,qu2024colt,anantha2023protip,zheng2024toolrerank}. 
In this approach, each tool’s description is converted into an embedding vector using an embedding model, and tools with the highest similarity to the user's query are then selected and integrated into the LLM's context.
However, existing retrieval methods exhibit two major limitations:
 (1) There is a noticeable semantic gap between tool descriptions and user queries, leading to inaccuracies in retrieval when embeddings are computed based on descriptions (Figure.~\ref{fig:tsne_visualization});
(2) Relying solely on embeddings for retrieval lacks scalability with an increasing number of tools as embeddings often lack expressiveness and fail to capture subtle nuances.

To address the distributional gap between query and tool embeddings and enhance retrieval accuracy, we introduce \OURS, a \textit{usage-driven} tool embedding generation method. 
Unlike traditional approaches that rely on tool descriptions, \OURS leverages example user queries associated with each tool to generate more accurate embeddings. 
Additionally, we propose a \textit{two-stage} tool retrieval method. 
In the first stage, the tool space is efficiently pruned, reducing the number of potential candidate tools. 
The second stage, which we call \RERANKER, refines the remaining tools to produce a more accurate final set of retrieved tools. 
\RERANKER is an accurate classification model that refines the output of the first stage by incorporating inter-tool and tool-query interactions, thereby capturing subtle nuances and improving retrieval accuracy.

One remaining challenge in training effective retrieval and refiner models is the need for domain-specific data. To overcome this, we build on insights from previous work~\cite{tiny-agent} which demonstrated that high-quality tool retrieval data can be generated using  LLMs~\cite{chen2024alpagasus,lee2024llm2llm,wei2024instructiongpt,cao2023instructionminingdatamining}.
Leveraging the powerful synthetic data generation capabilities of LLMs~\cite{openai2024gpt4technicalreport,llama3modelcard,jiang2024mixtral}, we additionally introduce \OURDATA, 
a comprehensive tool retrieval dataset specifically designed to train and evaluate retrieval systems across various domains. 
Our particular focus is on enhancing the natural co-occurrence of multiple tools and improving the naturalness of queries, ensuring that \OURDATA serves as a robust foundation for developing more effective tool retrieval systems.

\vspace{-4mm}
\paragraph{Contributions.}
In summary,  we make the following contributions to enhance tool retrieval performance:

\begin{itemize}[leftmargin=4mm]
\vspace{-2mm}
    \item We introduce \OURDATA, a new, high-quality domain-specific dataset for tool retrieval and instantiate three new datasets within this framework. When evaluated for quality by GPT-4-turbo, these datasets achieve a 60\% win rate compared to queries from ToolBench (\autoref{sec:tool_retrieval_dataset}).
    \vspace{-2mm}
    \item We propose \OURS, \textit{usage-driven} tool retrieval, as opposed to description-based tool retrieval of prior approaches. 
    Additionally, we introduce a \textit{two-stage} tool retrieval method which iteratively improves the quality of retrieved tools based on the retrieve-then-refine scheme (\autoref{sec:approach}).
    \vspace{-2mm}
    \item On the hardest ToolBench split, our method achieves over 25\% higher recall compared to ToolBench's retriever. Additionally, on our domain-specific datasets, our methods outperform description-based retrieval by over 30\% higher recall (\autoref{sec:experiments}).
\end{itemize}
\section{Related Work}
\subsection{Function Calling and Tool Use}
\label{sec:function-calling}
Function calling allows LLMs to interact with the world and agentic environments by filling in parameters to API functions and other tools. 
Typically, function descriptions and signatures are provided in the model's context window.
For accurate function calling, models must be able to choose the proper functions for the task and be able to fill in the correct parameters to those functions. 
Large models such as GPT-4 have demonstrated impressive function calling capabilities~\cite{kim2023llmcompiler}.
However, smaller models~\cite{patil2023gorilla, srinivasan2023nexusraven}, such as 7B and 13B models, have also been developed specifically for function calling tasks. The ToolBench~\cite{qin2024toolllm} dataset is a popular function calling dataset consisting of real-world APIs that was used to fine-tune a 7B LLaMA model for tool use.

\subsection{Tool Retrieval}
As discussed in~\autoref{sec:function-calling}, function descriptions and signatures are provided in the model's context window for applications relying on function calling. 
However, real-world applications often have hundreds or thousands of tools~\cite{qin2024toolllm}.
Providing information about all tools to the model may not be possible due to context length limits.
Furthermore, even when using models with longer context windows, providing all the tools in the prompt leads to significant compute and memory overheads~\cite{kim2023full}.
To address this, various tool retrieval methods have been proposed to select and provide only the relevant tools for incoming user queries instead of providing them all.

A notable approach to enhance tool retrieval performance is leveraging another LLM. 
AnyTool~\cite{du2024anytool} proposes to use GPT-4 for API retrieval and to further enhance retrieval performance through an iterative self-reflection method. 
Similarly, \cite{xu2024enhancing} incorporates a refiner LLM that iteratively refines user queries to boost retrieval performance. 
However, using LLMs for tool retrieval, along with iterative invocation, results in significant latency overhead of up to several seconds~\cite{xu2024enhancing}, limiting their use in various real-time applications.

Dense retrieval methods offer an efficient alternative, where each tool's description is embedded using an embedding model, and tools with the highest similarity to the embedding of the incoming user query are retrieved~\cite{qin2024toolllm}. 
ProTIP~\cite{anantha2023protip} adapts a dense retrieval model for iterative multi-tool selection.
ToolkenGPT~\cite{hao2024toolkengpt} proposes to learn an embedding of each tool that can be immediately used as an input token to LLMs.
COLT~\cite{qu2024colt} improves tool retrieval performance 
by fine-tuning the pre-trained encoder model through four distinct stages: semantic learning, collaborative learning, list-wise learning, and contrastive learning.

\OURS provides a different view of tool retrieval, which is tool embedding generated based on usage. It uses the user query embedding instead of tool description embedding to generate tool embeddings for retrieval.
A notable work is EasyTool~\cite{yuan2024easytool} which enhances tool leverages LLMs to rewrite tool descriptions, reducing inconsistency, redundancy, and incompleteness, ultimately improving retrieval performance. 
While EasyTool also proposes LLMs generate usage examples, these serve to provide in-context examples rather than directly improving the performance of the retriever models as in our work.

Another notable work is ToolRerank~\cite{zheng2024toolrerank}, an adaptive and hierarchy-aware reranking method for tool retrieval. 
\RERANKER differs from ToolRerank in several key ways. 
First, \RERANKER does not assume any hierarchy of tools. 
Second, \RERANKER processes all candidate tools retrieved from the first stage in one forward pass, whereas ToolRerank compares each tool description to the query one by one, which introduces latency overhead. Finally, \RERANKER utilizes \OURS embeddings as tool representations, while ToolRerank relies on tool descriptions.

\section{Dataset Generation}
\label{sec:tool_retrieval_dataset}
\begin{figure}[!t]
\centering
\begin{minipage}[t]{0.48\textwidth}
    \centering
    \includegraphics[width=.95\linewidth]{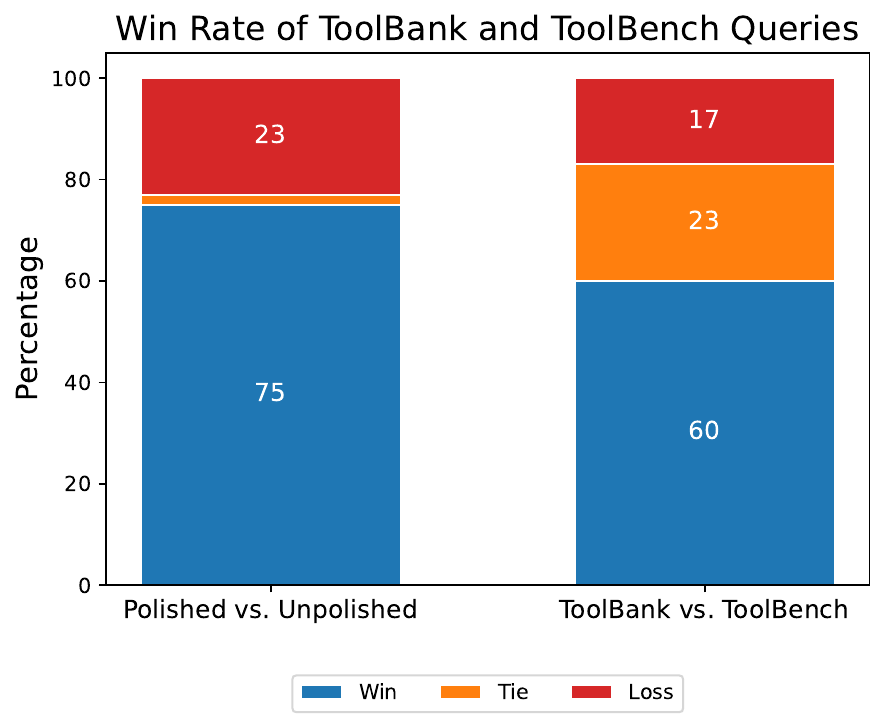}
    \caption{Comparison of naturalness, fluency, and coherence of queries. We first compare polished and unpolished queries within \OURDATA, with blue/orange/red bars indicating the number of times polished queries won, tied, or lost. Then, we compare queries from \OURDATA to those from ToolBench, using the same color scheme to represent the outcomes.}
    \label{fig:win_rate}
\end{minipage}
\hfill
\begin{minipage}[t]{0.48\textwidth}
    \centering
    \includegraphics[width=0.6\linewidth]{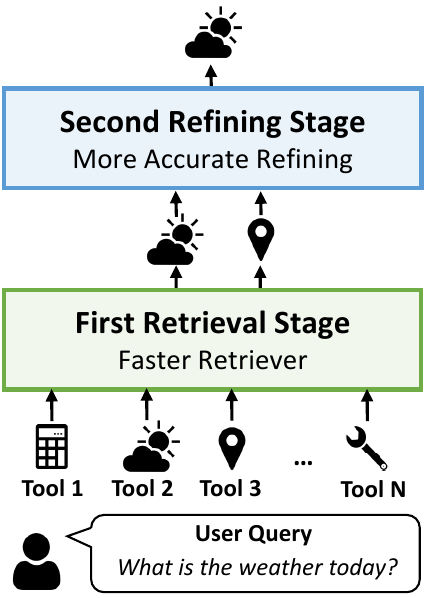}
    \caption{Illustration of two-stage tool retrieval. In the first stage, a fast retriever is used to prune the majority of the tools. In the second stage, a more accurate model is used to refine the tools kept in the first stage to get the final set of tools.}
    \label{fig:staged_retrieval}
\end{minipage}
\end{figure}

We generate domain-specific tool retrieval datasets with the goals of (1) demonstrating that users can create sufficiently large domain-specific datasets powered by LLMs~\cite{lee2024llm2llm,chen2024alpagasus,tiny-agent} for small tool retrieval models, 
and (2) addressing the limitations inherent in existing benchmarks~\cite{qin2024toolllm,chen2023t,xu2024enhancing,li2023api,huang2024metatool,tang2023toolalpaca}, 
which often lack coherent tool integration and query naturalness.

Particularly for the second aspect, current benchmarks frequently pair tools without considering their natural co-occurrence, leading to impractical and inconsistent combinations~\cite{qin2024toolllm,chen2023t,xu2024enhancing,li2023api,huang2024metatool,tang2023toolalpaca}. 
For example, a query from ToolBench— 
``Search for the companies that have been modified recently and fetch the lyrics for the song `Bad' by Michael Jackson'' pairs the
$\texttt{360 Business Tool}$ tool with the 
$\texttt{Chart Lyrics}$ tool, reflecting a clear mismatch in tool relevance.
This is because ToolBench randomly samples multiple tools from the tool pool, without much consideration of their co-ocurrance.

Moreover, due to the pairing of irrelevant tools, these benchmarks tend to be overly structured and verbose, resembling step-by-step queries rather than the more fluid, natural language typically used in real-world scenarios. 
For instance, a query in ToolBench dataset such as 
``Please provide me with details of breweries that are dog-friendly and have a patio, and include race details for race ID 207660, covering horses, jockeys, trainers, and their positions,'' showcases an unnatural pairing of unrelated tools, driven by a rigid, instructional style.

To this end, we curate a coherent and natural tool retrieval dataset \OURDATA that addresses limitations of existing benchmarks.
We aim to create the tool retrieval dataset that respects the natural co-occurrence of tools while ensuring more natural, real-world query queries.
To do so, we design the dataset generation process, which consists of the following two stages:


\begin{itemize}[leftmargin=2mm]
    \item \textbf{Query Generation}: 
    In this stage, we first sample \( T \) tools randomly from the entire tool set.
    In contrast to previous works where LLMs are prompted to use all \( T \) tools to generate an query~\cite{qin2024toolllm,chen2023t,xu2024enhancing,du2024anytool}, we allow them to select \( M \) tools, where \( M < T \), 
    that are coherent and contextually aligned.
    This approach promotes the natural co-occurrence of tools.
    We used \( T=10 \) and $M \in [2, 5]$ throughout our generation process, where we found sufficiently large 
    $T$ critical for LLMs to select tools that align contextually.
    Additionally, we provided 5 randomly sampled in-context examples to enhance generation quality and diversity.
    
    \item \textbf{Query Polish}: 
    Despite our query generation process improving tool co-occurrence, LLMs often produce step-by-step queries that seem unnatural. 
    To address this, we introduce an additional step to polish these initial, often robotic queries into fluent and concise English that more closely mirrors user queries in natural settings. 
\end{itemize}
\begin{figure}[!t]
    \centering
    \captionsetup{font=small}
    \includegraphics[width=0.8\linewidth]{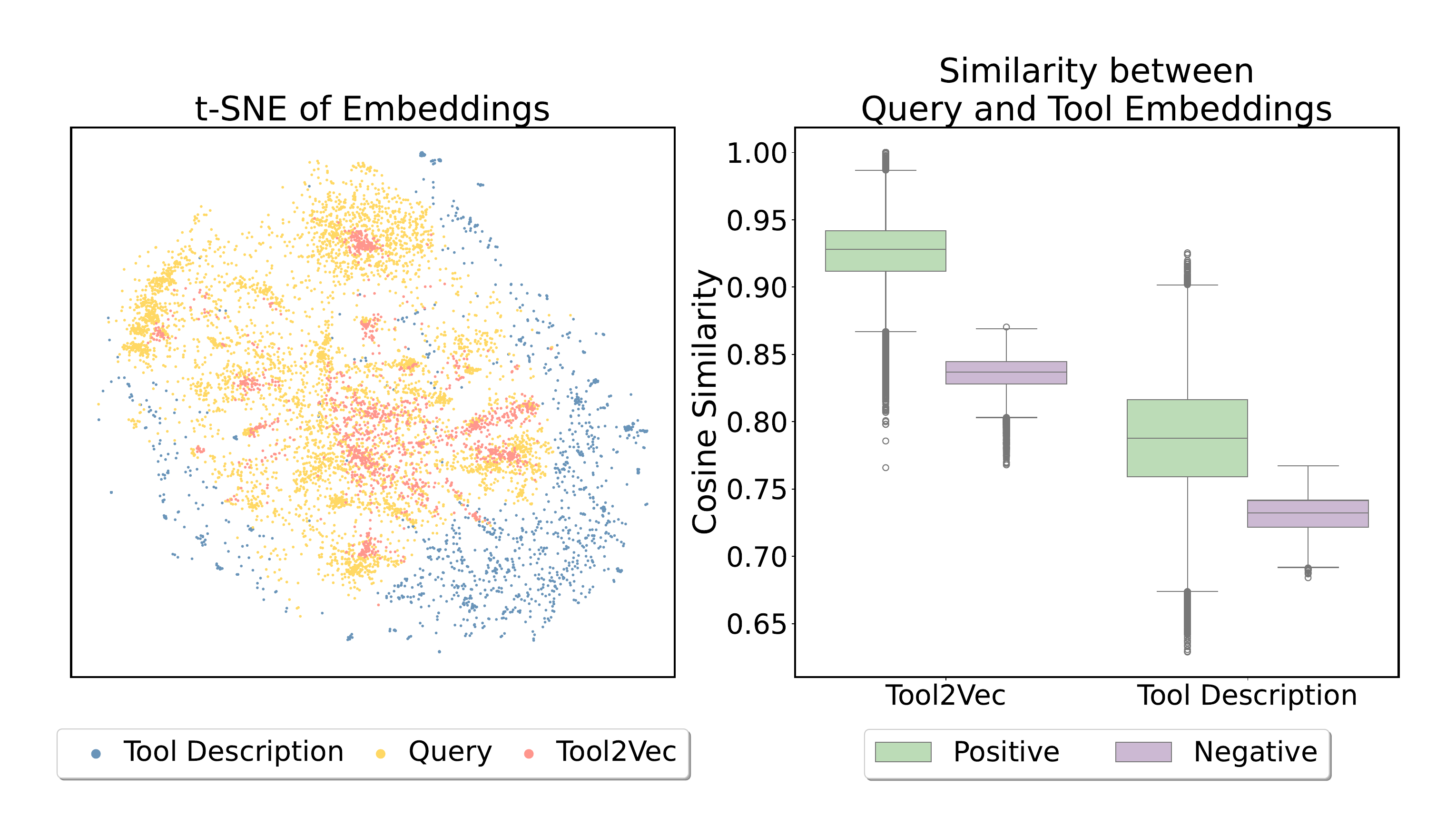}
    \caption{(Left) t-SNE visualization of embeddings for queries, \OURS, and tool descriptions. (Right) Cosine similarity between instruction and tool embeddings. 
    The figure displays two distributions for both \OURS embeddings and tool description embeddings: one labeled `Positive,' representing cosine similarity between queries and the embeddings of tools used for those instructions, 
    and the other labeled `Negative,' representing cosine similarity between instructions 
    and the embeddings of tools not used for those queries.} 
\label{fig:tsne_visualization}
\end{figure}

We compare \OURDATA against ToolBench~\cite{qin2024toolllm}, one of the most widely adopted benchmarks for tool retrieval in \autoref{fig:win_rate}.
The study illustrated in \autoref{fig:win_rate} directly evaluates the naturalness, fluency, and coherence of queries.
We randomly sample 100 queries from both the unpolished and polished versions of \OURDATA, 
as well as 100 queries from ToolBench. 
GPT-4-turbo is then tasked with judging which queries are superior based on the aforementioned criteria~\cite{zheng2024judging}. 
The results demonstrate that Query Polish consistently generates queries that outscore both the baseline unpolished queries and the queries from the ToolBench dataset.
We present additional analyses, including a qualitative comparison between \OURDATA and ToolBench in \autoref{fig:qualitative_analysis}, 
as well as the impact of the Query Polish step on the quality of synthetic queries in \OURDATA, shown in \autoref{fig:unpolished_instructions}. 
A more detailed explanation is provided in \autoref{appendix:dataset_generation_details}.

\section{Tool Retrieval Approaches}
\label{sec:approach}

In this section, we propose a two-stage tool retrieval method, as outlined in \autoref{fig:staged_retrieval}. 
The first \textit{retrieval} stage efficiently prunes the majority of the tool space, while the second \textit{refining} stage further refines the kept tools to produce the final set of retrieved tools. 
This is analogous to retrieve-then-rerank pipelines for document retrieval~\cite{nogueira2019passage}. 

The benefit of this two-stage approach is that the more powerful refining stage can correct any errors from the fast retrieval stage. 
Additionally, because retrieval is performed first to filter out candidate tools, the refiner stage does not need to operate on the entire toolset, leading to efficient implementation and runtime.
For the first stage, we present two methods:
(i) usage-driven embedding generation (\autoref{subsec:t2v}) and
(ii) multi-label classification (\autoref{subsec:mlc}).
For the second stage, we introduce an efficient yet accurate classification model for refining the output of the first stage (\autoref{subsec:reranker}), which can contextualize the inter-tool and tool-query interactions.

\begin{figure*}[!t]
    \centering
    \captionsetup{font=small}
    \includegraphics[width=\linewidth]{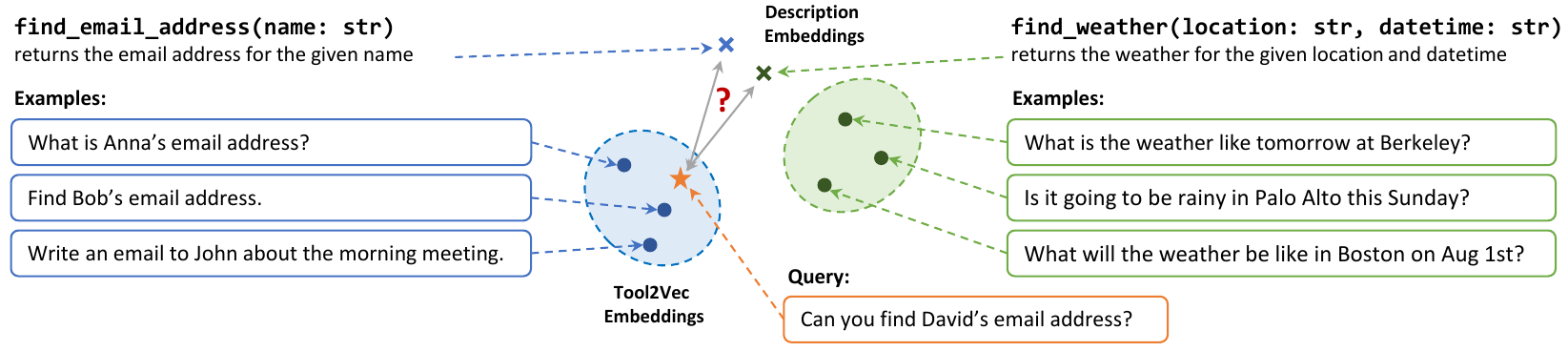}
    \caption{Illustration of how user query embeddings are used as tool embeddings. 
    The embeddings of example queries in the left side of figure corresponds to the tool \texttt{find\_email\_address} \OURS embedding. 
    If multiple queries use the same tool, their embeddings are averaged. 
    Likewise, the \OURS embedding of \texttt{find\_weather} is the average of the embeddings from the examples shown on the right side of the figure. 
    The disjoint embedding distributions reflect the different semantics of the two sets of examples.
    However, the description embeddings of those two tools are not close to each cluster because of the semantic domain gap between query and tool description, which leads to the suboptimal retrieval performance. } 
    \label{fig:t2v}
\end{figure*}

\subsection{\OURS: Usage-Driven Embedding Generation}
\label{subsec:t2v}
Similar to document retrieval, performing vector search over embeddings can be used to retrieve an initial set of tools in the first stage.
Previous tool retrieval methods have relied on tool descriptions to obtain embeddings of each tool~\cite{anantha2023protip,qin2024toolllm,qu2024colt,yuan2024easytool}.
However, this approach may be suboptimal due to the semantic disparity between tool descriptions and user queries.
\autoref{fig:tsne_visualization} (Left) illustrates how tool descriptions and user queries can be disjoint in the embedding space, making tool retrieval based on embedding similarity challenging.
This issue persists even when the descriptions are augmented with additional information, such as tool code, to improve retrieval performance~\cite{yuan2024easytool,zheng2024toolrerank,du2024anytool}.

To reduce the distributional gap between query and tool embeddings for retrieval, 
we propose \OURS, the usage-driven tool embedding generation. 
Instead of using tool descriptions, we propose to use \textit{user queries} to obtain tool embeddings. 
In more detail,
if we have multiple user queries that use a specific tool, we use the average embeddings of those user queries as the \OURS embedding that represents the tool.
For example, in \autoref{fig:t2v}, 
we have multiple user queries that use the tool \texttt{find\_email\_address}, such as ``What is Anna’s email address?" 
In this case, we use an embedding model (e.g., E5~\cite{Wang2022TextEB}) to obtain the embedding for each user query, and the average of these embeddings is used as the \OURS embedding for the tool \texttt{find\_email\_address}.
Likewise, the \OURS embedding for the tool \texttt{find\_weather} can be obtained the same way using the associated user queries.
As shown in the figure, 
since the \OURS embeddings of these tools are derived from user queries, 
they are closer to the incoming user query in the embedding space compared to embeddings derived from tool descriptions.

\begin{figure*}[!t]
    \centering
    \captionsetup{font=small}
    \includegraphics[width=0.9\linewidth]{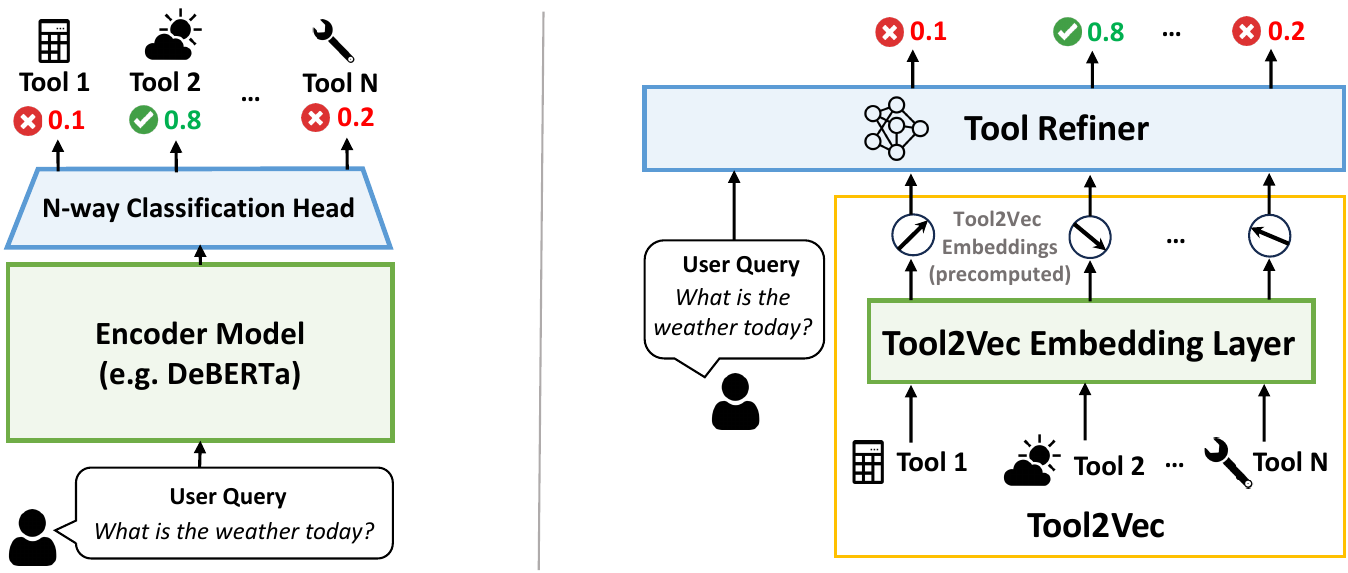}
    \caption{(Left) Illustration of \SIDFORMER: 
    The encoder model (e.g., DeBERTa~\cite{he2023debertav}) takes user query tokens as input and outputs the probability of each tool.
    (Right) Illustration of \RERANKER: 
    The fine-tuned encoder model takes the user query and \OURS embeddings of retrieved tools as inputs. 
    We precompute the \OURS embeddings and use them in conjunction with the user query. 
    The pre-trained encoder model is then fine-tuned with binary classification loss for each tool.
} 
    \label{fig:toolrefiner}
\end{figure*}

To further justify the benefits of \OURS's usage-driven tool embedding generation, we perform an analysis as illustrated in \autoref{fig:tsne_visualization}.
The left figure is a t-SNE visualization of the embeddings of the user queries, \OURS, and tool descriptions. 
It shows that the query embeddings form clusters, with \OURS embeddings typically positioned at the centroids of these clusters. 
The tool description embeddings, however, are scattered outside of the distributions of instruction embeddings. 
Evidently, this is due to the semantic gap between the tool description and user query.

The right figure is the box plots with interquartile ranges (IQR) of the cosine similarity between the instruction and tool embeddings.
It shows two distributions: 
`Positive' for the similarity between instruction embeddings and the embeddings of tools used to process the given instructions, 
and `Negative' for the similarity between instruction embeddings and the embeddings of tools not used.
For \OURS embeddings, the positive and negative distributions do not overlap, 
indicating a clear distinction. 
However, the cosine similarity distributions for tool descriptions show significant overlap between positive and negative, 
implying that the traditional tool description embeddings are less effective at distinguishing between relevant and irrelevant tools compared to the \OURS embeddings.

\subsection{Tool Retrieval as Multi-Label Classification}
\label{subsec:mlc}
In settings where there are enough instructions and associated tool labels, the first stage of our two-stage tool retrieval (\autoref{fig:staged_retrieval}) can alternatively be formulated as a multi-label classification problem.
Furthermore, given the rise of synthetic data generation methods~\cite{chen2024alpagasus,lee2024llm2llm,wei2024instructiongpt,cao2023instructionminingdatamining}, 
it has become possible to construct such labeled high-quality pairs synthetically with the competent LLMs~\cite{llama3modelcard,openai2024gpt4technicalreport,jiang2023mistral,jiang2024mixtralexperts}, 
as demonstrated in \autoref{sec:tool_retrieval_dataset}.

One approach for multi-label classification (\SIDFORMER) involves training a model that takes the instruction as input and outputs the classification logits for each tool, as illustrated in the left figure of \autoref{fig:toolrefiner}. 
When a user query is provided, such as ``What is the weather today?", we assign a label of 1 to all required tools and a label of 0 to unused tools to training the \SIDFORMER.
In this example, the $\texttt{find\_weather}$ tool receives a label of 1, while other tools receive a label of 0. 
To achieve this, we fine-tune the pre-trained DeBERTa-V3 base model~\cite{he2023debertav,devlin-etal-2019-bert}, which features a $H\times T$ classification head operating on the output [CLS] token. 
Here, $H$ represents the dimension of the [CLS] token, and $T$ denotes the total number of tools in the dataset.
Surprisingly, this simple \SIDFORMER alone demonstrates strong tool retrieval capabilities, even surpassing other state-of-the-art methods, as discussed in \autoref{subsec:toolbench_results}.

\subsection{\RERANKER}
\label{subsec:reranker}
As the number of tools increases, methods like \OURS and \SIDFORMER, though effective, may struggle with the vast set of tools. 
This is because these methods can miss relevant tools due to unnecessary noise between tools.
Additionally, \OURS embeddings and \SIDFORMER cannot capture the tool-query and tool-tool interactions, which limits their effectiveness. 

To address these challenges, we introduce \RERANKER for the second stage tool retrieval.
\RERANKER enhances tool retrieval performance on top of any initial tool retrieval method, such as those outlined in \autoref{subsec:t2v} and \autoref{subsec:mlc}.
Once an initial set of tools are retrieved in the first stage, \RERANKER then further classifies whether the retrieved tools are relevant or not.
As illustrated in \autoref{fig:toolrefiner}, 
\RERANKER takes the user query and the \OURS embeddings of tools retrieved in the first stage to classify which tools are needded to process the user query.
We fine-tune the pre-trained DeBERTa-V3~\cite{he2023debertav} xsmall model to get \RERANKER.
Similar to \SIDFORMER, we assign a label of 1 to all required tools and a label of 0 to unused tools. 
For instance, if the user query is ``What is the weather today?", 
the $\texttt{find\_weather}$ tool receives a label of 1, while other tools receive a label of 0. 
We then calculate the binary cross-entropy loss at each tool position. 
By doing so, \RERANKER understands the interactions between query and tool and therefore improves the retrieval performance.
Notably, if the \OURS embedding is pre-computed, \RERANKER can be used on top of any other retrieval methods to improve the performance.

This approach is analogous to passage reranking~\cite{nogueira2019passage, yilmaz2019applying},
which determines the ranking of retrieved documents based on their similarity to the query. Similar to passage reranking, the \OURS is trained as a classification model. The key difference is that while traditional passage rerankers evaluate and order the similarities of retrieved documents one by one in relation to the query, \RERANKER simultaneously reranks all retrieved tools. Furthermore, the reranker operates directly on Tool2Vec embeddings.

\begin{table*}[!t]
\caption{
Comparison of tool retrieval results on the ToolBench dataset. We compared our methods against two baselines: the ToolBench retriever~\cite{qin2024toolllm} and COLT~\cite{qu2024colt}. Evaluation metrics include Recall@K, where K values are 3, 5, and 7. In the table, R@K stands for Recall@K. The best-performing method is highlighted in boldface, while the second-best performing method is underlined. We reproduce the ToolBench retriever results based on the original codebase. For the other baseline method, COLT, we report the numbers available in the paper~\cite{qu2024colt}. 
We observe similar trends when evaluating nDCG@K, as shown in \autoref{table:toolbench_ndcg}.
}
\begin{center}
\footnotesize{
\setlength{\tabcolsep}{6pt}{
\begin{tabular}{c|c|c|c|c|c|c|c|c|c}
\toprule
  \multirow{2}{*}{Method}& \multicolumn{3}{c|}{\textbf{ToolBench I1}} & \multicolumn{3}{c|}{\textbf{ToolBench I2}} & \multicolumn{3}{c}{\textbf{ToolBench I3}} \\
 & R@3 & R@5 & R@7 & R@3 & R@5 & R@7 & R@3 & R@5 & R@7 \\
 \cmidrule{1-10}
 ToolBench Retriever & 79.97 & 90.19 & 93.21 & 67.25 & 78.25 & 85.75 & 54.07 & 63.88 & 73.73 \\
 COLT & - & - & - & 75.72 & 85.03 & - & 76.63 & 85.50 & - \\
 \cmidrule{1-10}
 \highlightrow\OURS & 85.88 & 93.29 & 94.42 & 72.79 & 79.67 & 82.75 & 75.23 & 84.90 & 86.60 \\
 \highlightrow\SIDFORMER & \underline{91.80} & \underline{96.00} & \underline{96.67} & \underline{80.67} & \underline{85.63} & \underline{87.46} & \textbf{81.35} & \graycell{86.27} & 88.27  \\
 
 \highlightrowb\RERANKER + \OURS & 89.63 & 95.33 & 96.17 & 76.83 & 84.42 & 86.38 & \underline{80.58} & \textbf{87.80} & \textbf{89.70} \\
 \highlightrowb\RERANKER + \SIDFORMER  & \textbf{91.84} & \textbf{96.83} & \textbf{97.01} & \textbf{82.89} & \textbf{87.92} & \textbf{88.96} & 79.83 & \underline{86.91} & \underline{88.98}   \\
\bottomrule
\end{tabular}
}
}
\end{center}
\vspace{-4mm}
\label{table:toolbench_results}
\end{table*}
\section{Experiments}
\label{sec:experiments}
In this section, we describe experimental results that validate the effectiveness of our proposed methods, \OURS, \SIDFORMER, 
and \RERANKER on various benchmarks including 
ToolBench~\cite{qin2024toolllm} and \OURDATA.

\subsection{ToolBench}
\label{subsec:toolbench_results}
In \autoref{table:toolbench_results}, we evaluate our proposed methods in the ToolBench dataset~\cite{qin2024toolllm}, comparing their performance against two established baselines: the ToolBench Retriever~\cite{qin2024toolllm}, and COLT~\cite{qu2024colt}. 
We observe that our methods constantly outperform the baselines with large margins.

\subsubsection{Experimental Details}
For benchmarking, we use the ToolBench dataset, which is the current standard benchmark for multi-tool retrieval.
The dataset is divided into three subsets (I1, I2, and I3), and each subset corresponds to different levels in the RapidAPI Hub tool hierarchy. 
As the subset number increases from I1 to I3, the tools used are sampled from higher levels of the hierarchy. 
This means that I3 involves more complex or broadly categorized tools compared to I1 and I2.

For all methods used in these experiments, pre-trained encoder transformer models are fine-tuned to each subset of the dataset. 
The COLT retriever, on the other hand, is a fine-tuned version of Contriever~\cite{izacard2022unsupervised}, another dense retrieval model based on BERT-base.

For our methods, \SIDFORMER and \RERANKER, we use DeBERTaV3~\cite{he2023debertav}. Specifically, \SIDFORMER uses DeBERTaV3-base (86 million parameters) and \RERANKER uses DeBERTaV3-xsmall (22 million parameters). 
To get \OURS embedding, we fine-tune pre-trained E5-base~\cite{Wang2022TextEB} model. The model is fine-tuned with triplet loss for one epoch.

\subsubsection{Result Analysis}
\autoref{table:toolbench_results} presents the performance comparison. The first two rows show the baseline methods: ToolBench retriever~\cite{qin2024toolllm} and COLT retriever~\cite{qu2024colt}. 
The last four rows display our methods. The first two rows represent the first-stage fast retrieval methods: \OURS and \SIDFORMER. 
The last two rows present the results for the two-stage methods: \RERANKER combined with \OURS, where the first stage of retrieval is performed by \OURS, and \RERANKER combined with \SIDFORMER, where the first stage of retrieval is performed by \SIDFORMER.

The results are summarized in \autoref{table:toolbench_results}. We use Recall@K as the primary evaluation metric, with K values of 3, 5, and 7.
nDCG results are provided in \autoref{appendix:ndcg_results}.
The nDCG values in \autoref{table:toolbench_ndcg} also exhibit a similar trend to the Recall results presented in \autoref{table:toolbench_results}.
The results for the ToolBench retriever are reproduced using the original codebase, while the Recall values for COLT are taken from~\cite{qu2024colt}, as the codebase is unavailable for reproduction.

\SIDFORMER and \RERANKER consistently outperform the baseline methods by significant margins across all ToolBench subsets. 
\OURS outperforms the ToolBench retriever across all subsets but falls short of the COLT retriever. 
Comparing the third and fifth rows in \autoref{table:toolbench_results}, \RERANKER achieves up to 3.8 additional Recall@K across all subsets. 
For \SIDFORMER, \RERANKER shows improvements of up to 2.3 Recall@K for subsets I1 and I2.


\begin{table*}[!t]
\vspace{3mm}
\caption{
We compare tool retrieval outcomes using the \OURDATA dataset. 
The baseline consists of methods that identify tools based on their descriptions. 
We evaluate performance using the evaluation metric Recall@K for K values of 3, 5, and 7. 
The results are organized into three sections: the first three columns show outcomes using \NUMPY, 
the following three columns display the results with \PANDAS, 
and the final three columns present the results for \AWS.
We present the E5-base results fine-tuned with the tool description as the baseline. 
The best-performing method is highlighted in boldface, while the second-best performing method is underlined.
} 
\vspace{-1mm}
\begin{center}
\footnotesize{
\setlength{\tabcolsep}{6pt}{
\begin{tabular}{c|c|c|c|c|c|c|c|c|c}
\toprule
  \multirow{2}{*}{Method}& \multicolumn{3}{c|}{\textbf{\NUMPY}} & \multicolumn{3}{c|}{\textbf{\PANDAS}} & \multicolumn{3}{c}{\textbf{\AWS}} \\
 & R@3 & R@5 & R@7 & R@3 & R@5 & R@7 & R@3 & R@5 & R@7 \\
 \cmidrule{1-10}
Description-Based Retriever & 50.82 & 64.09 & 71.84 & 27.86 & 34.90 & 40.00 & 41.92 & 46.46 & 49.13 \\
 \cmidrule{1-10}
  \highlightrow \OURS & 52.97 & 64.18 & 71.11 & 36.52 & 42.01 & 45.17 & 55.38 & 63.14 & 67.98 \\
 \highlightrow \SIDFORMER & 70.35 & \underline{80.78} & \underline{84.73} & 41.49 & \underline{49.69} & \underline{54.34} & \underline{70.99} & \underline{79.69} & \underline{82.91} \\
 \highlightrowb \RERANKER + \OURS & \underline{71.61} & 79.52 & 82.22 & \underline{42.94} & 47.65 & 49.33 & 69.12 & 74.08 & 75.43 \\
 \highlightrowb \RERANKER + \SIDFORMER & \textbf{73.82} & \textbf{84.24} & \textbf{87.47} & \textbf{47.76} & \textbf{55.28} & \textbf{59.13} & \textbf{72.42} & \textbf{81.17} & \textbf{84.49} \\
\bottomrule
\end{tabular}
}
}
\end{center}
\vspace{-4mm}
\label{table:synthetic_resutls}
\end{table*}

\subsection{\OURDATA} 
In this section, we benchmark the methods introduced in \autoref{sec:approach}
with our new dataset, \OURDATA. 
The results are summarized in \autoref{table:synthetic_resutls}.
The baseline is a description based retrieval method. 
For \OURDATA, we observe that our methods outperform the baselines, following similar trends to those in \autoref{subsec:toolbench_results}.

\subsubsection{Experimental Details}
The baseline used in this experiment is E5-base model~\cite{Wang2022TextEB}, fine-tuned with the description of tools. We conduct an extensive hyperparameter search for the baselines to rigorously evaluate our methods.
Similar to \autoref{subsec:toolbench_results}, we fine-tune pre-trained encoder model for \SIDFORMER, \OURS, and \RERANKER.

For all subsets in this data, we split the training set into an 8:2 ratio for training and validation. 
We conduct hyperparameter tuning using the validation set and report performance on the test set using the best-performing hyperparameters. 
We evaluate the test set only once across all experiments.

\subsubsection{Result Analysis}
In \autoref{table:synthetic_resutls}, the first row is the result with the description-based baseline. 
Other rows are results with our methods. 
The first two rows represent the first-stage fast retrieval methods: \OURS and \SIDFORMER. 
The last two rows present the results for the two-stage methods: \RERANKER combined with \OURS, where the first stage of retrieval is performed by \OURS, and \RERANKER combined with \SIDFORMER, where the first stage of retrieval is performed by \SIDFORMER.
We also use the Recall@K as the primary metric for evaluations in this experiment.

All of our methods outperform the baseline by up to 30 additional Recall@K.
We observe that \RERANKER improves the retrieval results consistently for both \OURS and \SIDFORMER.
Especially, the improvement is remarkable when \RERANKER is used with \OURS, which have the gain up to 21 for Recall@K. 

We observe that our models perform worse on the Pandas dataset; specifically, the \RERANKER combined with \SIDFORMER achieves 25\% less Recall@3 on \PANDAS dataset than both the \NUMPY and \AWS datasets. \PANDAS dataset contains various data types like time series, periods, intervals, and indexes; hence, the model is mostly confused about which data type to operate on. 
For the further detail, please refer to \autoref{appendix:pandas_worse}.
\section{Discussion}
\subsection{Ablation Studies}
We conduct several ablation studies. 
First, we study providing \RERANKER embeddings of tool descriptions, rather than \OURS embeddings.
As shown in \autoref{table:t2v_vs_tool_description}, we find that providing \OURS embeddings significantly improves the performance of \RERANKER.
Additionally, we examine the effect on \RERANKER performance when varying the number of retrieved tools in the first stage as provided in \autoref{table:num-candidate-tools}.
We observe that the \RERANKER performance improves at some point but decreases after some point.
Furthermore, we analyze the quality of retrieval with \OURS compared to description-based retrieval across a variety of embedding models. We do not perform any fine-tuning of the embedding models and use both open and closed source models of various sizes. As shown in \autoref{table:emb-model-ablation}, \OURS leads to significant improvement in retrieval quality compared to description-based retrieval across all models.
For more details about above studies, please refer to \autoref{appendix:ablation_studies}.

\subsection{Analysis on How \RERANKER Improves Performance}
We analyze how \RERANKER improves the performance compared to first-stage retrieval.
We find \RERANKER excels in handling complex queries and maintains consistent performance across tools, while \OURS struggles more with simpler queries and shows higher error rates on certain tools. We provide more details in \autoref{appendix:discussion}. 
\section{Conclusions}
Although function calling allows LLM agents to perform a wider range of tasks beyond their inherent capabilities, as tasks become increasingly complex, the set of tools required can grow vast. 
This expansion can lead to context window limitations and system overheads caused by long prompts, which can also degrade performance. 
For this reason, we propose an efficient two-stage tool retrieval system that combines a fast first-stage tool retriever, \OURS and \SIDFORMER, with a powerful second-stage tool refiner, \RERANKER.
Additionally, domain-specific dataset generation is critical for building specialized tool retrieval applications. 
Current LLMs, however, can generate high-quality tool retrieval data.
To demonstrate this, we create a new tool retrieval dataset, \OURDATA with LLMs.
Our retrieval strategy achieves over 25\% higher Recall than ToolBench's description-based retriever and outperforms description-based retrieval by up to 30\% on \OURDATA.
We look forward to future work building upon our framework, including dataset generation and tool retrieval methods, to streamline tool-augmented LLMs for complex, large-scale tasks. 

\section*{Acknowledgements}
We acknowledge gracious support from Furiosa team including June Paik, Jihoon Yoon, and Hyung Koo.
We also appreciate the support from Microsoft through their Accelerating Foundation Model Research, including
great support from Sean Kuno and Dan Fay.
Furthermore, we appreciate support from
Google Cloud, the Google TRC team, and specifically Jonathan Caton, and Prof. David Patterson.
Prof. Keutzer's lab is sponsored by the Intel corporation, Intel One-API, Intel VLAB team, the Intel One-API center of
excellence, Apple, Samsung, Panasonic, Nvidia, as well as funding through BDD and BAIR.
We appreciate great feedback and support from Ellick Chan, Saurabh Tangri, Andres
Rodriguez, and Kittur Ganesh from Intel.
Sehoon Kim and Suhong Moon would like to acknowledge the support from the Korea Foundation for Advanced Studies (KFAS).
Our conclusions do not necessarily reflect the position or the policy of our sponsors, and no official endorsement should be~inferred.

\bibliographystyle{plain}
\bibliography{aaai25}

\appendix
\counterwithin{figure}{section}
\counterwithin{table}{section}
\clearpage
\onecolumn
\section{Additional Results}
\subsection{Ablation Studies}
\label{appendix:ablation_studies}
This section details the ablation studies. 
First, we investigate the effectiveness for \RERANKER performance of \OURS embeddings compared to tool description embeddings and find that \OURS consistently outperforms tool description embeddings. 
Then, we explore the impact of the number of initially retrieved candidate tools on overall retriever performance. 
We observe that increasing the number of candidate tools consistently enhances performance up to a certain point, 
after which the improvement plateaus and the retrieval metrics degrade.
Finally, we compare \OURS-based retrieval and description-based retrieval for various embedding models.
We find that \OURS-based retrieval outperforms description-based retrieval consistently over all embedding models.

\subsubsection{\OURS vs. Tool Description Embeddings}

In this ablation study, we demonstrate the effectiveness of using \OURS embeddings to the tool description embeddings for \RERANKER. 
Specifically, we train two \RERANKER models, one with \OURS embeddings and one with the tool description embeddings, on ToolBench I3 dataset.
We retrieve top-64 tools first by cosine similarites between \OURS embeddings and query embeddings.
We observe that \RERANKER trained with \OURS embeddings outperforms \RERANKER trained with tool description embeddings across most retrieval settings. 
Our results are shown in \autoref{table:t2v_vs_tool_description}.


\begin{table*}[!t]
\caption{Performance comparison of \RERANKER with \OURS embeddings and tool description embeddings on ToolBench I3. 
The first row represents \RERANKER fine-tuned with \OURS tool embeddings using \OURS-based retrieval, the second row represents \RERANKER fine-tuned with description embeddings using \OURS-based retrieval, and the third row represents \RERANKER fine-tuned with description embeddings using description based retrieval. For each row, we fine-tune the E5-base embedding model specifically for each use case to compute the embeddings.}
\vspace{-1mm}
\begin{center}
\footnotesize{
\setlength{\tabcolsep}{4pt}{
\begin{tabular}{l|l|ccc}
\toprule
\textbf{Embedding for \RERANKER} & \textbf{Retrieval Method} & \textbf{Recall @ 3} & \textbf{Recall @ 5} & \textbf{Recall @ 7} \\
\midrule
\highlightrowb \OURS & \OURS & 80.58 & 87.80 & 89.70 \\
Tool Description & \OURS & 71.55 & 82.27 & 87.28 \\
Tool Description & Tool Description & 66.00 & 74.60 & 76.55 \\
\bottomrule
\end{tabular}
}
}
\end{center}
\vspace{-4mm}
\label{table:t2v_vs_tool_description}
\end{table*}


\begin{table*}[!t]
\caption{Comparison of \RERANKER in \autoref{subsec:reranker} performance on the ToolBench dataset across multiple top-$N$ candidate tool configurations. 
We use an \SIDFORMER-based retriever and a \OURS-based retriever to retrieve a set of $N$ candidate tools where N varies from 8 to 128. 
Our evaluation metric is Recall@$K$, where $K$ are values 3, 5, and 7. 
The best-performing top-$N$ configuration for each retriever method is highlighted in boldface.}
\vspace{-1mm}
\begin{center}
\footnotesize{
\setlength{\tabcolsep}{4pt}{
\begin{tabular}{l|c|ccc|ccc|ccc}
\toprule
\multirow{2}{*}{Method} & \multirow{2}{*}{Top-$N$} & \multicolumn{3}{c|}{\textbf{ToolBench I1}} & \multicolumn{3}{c|}{\textbf{ToolBench I2}} & \multicolumn{3}{c}{\textbf{ToolBench I3}} \\
 & & R@3 & R@5 & R@7 & R@3 & R@5 & R@7 & R@3 & R@5 & R@7 \\
\midrule
\multirow{5}{*}{\SIDFORMER Retriever} 
& 8 & 91.18 & 95.17 & 96.33 & 81.96 & 87.54 & 88.21 & 70.53 & 82.90 & 86.63 \\
& 16 & 91.59 & 96.42 & \textbf{97.08} & 81.96 & 87.54 & 88.21 & 78.13 & 86.43 & 87.95 \\
& 32 & 91.43 & 96.25 & \textbf{97.08} & 81.67 & 87.33 & 88.58 & \textbf{79.83} & \textbf{86.81} & \textbf{88.98} \\
& 64 & \textbf{91.84} & \textbf{96.83} & 97.01 & \textbf{82.89} & \textbf{87.92} & \textbf{88.96} & 76.75 & 85.88 & 86.80 \\
& 128 & 90.67 & 96.25 & 96.67 & 80.17 & 87.17 & 89.12 & 77.08 & 85.72 & 87.98 \\
\cmidrule{1-11}
\multirow{5}{*}{\OURS Retriever} 
& 8 & 87.01 & 93.79 & 95.00 & 75.25 & 81.25 & 82.75 & 74.00 & 87.38 & 89.22 \\
& 16 & 89.76 & 94.79 & 94.96 & 77.96 & 83.21 & 84.46 & 74.00 & 87.38 & 89.22 \\
& 32 & \textbf{90.05} & 94.46 & 95.25 & 76.88 & 83.17 & 84.33 & 79.50 & 87.77 & 89.53 \\
& 64 & 89.63 & \textbf{95.33} & \textbf{96.17} & 76.83 & \textbf{84.42} & \textbf{86.38} & \textbf{80.58} & \textbf{87.80} & \textbf{89.70} \\
& 128 & 87.84 & 94.87 & 95.42 & \textbf{77.17} & 82.42 & 83.87 & 78.17 & 87.55 & 89.30 \\
\bottomrule
\end{tabular}
}
}
\end{center}
\vspace{-4mm}
\label{table:num-candidate-tools}
\end{table*}


\subsubsection{Analysis on the Impact of the Number of Candidate Tools}

In this set of experiments, we explore the impact of the number of candidate tools on the overall retrieval performance of \RERANKER. 
Specifically, for each query, we retrieve the top-$N$ tools either from the output of \SIDFORMER or from cosine-similarity-based retrieval between the user query embedding and the \OURS. Then, we fine-tune the pre-trained DeBERTa-v3 xsmall model with these $N$ tools. 
We vary the value of $N$ across 8, 16, 32, 64, and 128 and evaluate performance on the ToolBench dataset. 

In \autoref{table:num-candidate-tools}, one key observation is the initial improvement in performance as $N$ increases. 
This trend is consistent across all datasets and retrieval methods, 
but the performance improvement plateaus after a certain $N$ value, with peak performance achieved at $N$=32 or 64 configurations. 
Specifically, for ToolBench I1 and I2, the best-performing $N$ value is 64 for both retrieval methods, while for ToolBench I3, the best-performing $N$ value is 64 for the \OURS-based retriever and 32 for the \SIDFORMER-based retriever. 
This is because the retrieval performance improves as $N$ increases.
However, the performance of the \RERANKER method decreases for large $N$ across all datasets and retrieval methods, 
indicating that including too many candidate tools can overwhelm the language model and lead to confusion and suboptimal performance.

Moreover, comparing the performance of different retrieval methods, the \SIDFORMER-based retriever consistently outperforms the \OURS-based retriever for ToolBench I1 and I2 datasets across most of the top-$N$ settings, 
while the \OURS-based retriever outperforms the \SIDFORMER-based retriever for the ToolBench I3 dataset. 
This suggests that the choice of retrieval method can significantly impact the performance of the \RERANKER method, and the optimal $N$ value may vary depending on the dataset and retrieval method used.

From these observations, we can conclude that it is critical to carefully select the appropriate $N$ value when training a tool retriever.
While lower $N$ values enable faster inference, 
they may result in worse performance when dealing with a large number of tools. 
Conversely, including too many candidate tools can confuse \RERANKER, leading to worse performance than the performance with smaller $N$. 
This indicates the importance of balancing the trade-off between performance and efficiency when designing a tool retriever for a given dataset.

\subsubsection{Various Embedding Models}
The experiments performed in \autoref{sec:experiments} rely on E5-base as an embedding model. To demonstrate the effectiveness of Tool2Vec compared to description-based retrieval, we show the results with other embedding models in \autoref{table:emb-model-ablation}. Tool2Vec consistently outperforms description-based retrieval across model families and sizes.

\begin{table*}[h]
\caption{Comparison of Tool2Vec retrieval and description-based retrieval across various embedding models on ToolBench's I3 split. Models are evaluated without any fine-tuning. Tool2Vec consistently outperforms description-based retrieval on both open source and closed embedding models.}
\begin{center}
\footnotesize{
\setlength{\tabcolsep}{4pt}{
\begin{tabular}{l|ccccc}
\toprule
Method & R@3 & R@5 & R@7 & R@10 & R@12 \\
\midrule
\highlightrowb E5-small + Tool2Vec & 63.12 & 75.95 & 82.27 & 85.87 & 86.73 \\
E5-small + Descriptions & 20.62 & 30.45 & 37.27 & 42.42 & 46.87 \\
\midrule
\highlightrowb E5-base + Tool2Vec & 62.48 & 76.10 & 80.17 & 84.80 & 86.45 \\
E5-base + Descriptions & 32.12 & 38.97 & 43.92 & 50.63 & 54.80 \\
\midrule
\highlightrowb E5-large + Tool2Vec & 60.40 & 71.20 & 77.92 & 84.18 & 85.93 \\
E5-large + Descriptions & 33.28 & 42.12 & 48.45 & 56.48 & 60.25 \\
\midrule
\highlightrowb Mxbai-embed-large + Tool2Vec & 59.23 & 67.97 & 76.30 & 80.78 & 83.68 \\
Mxbai-embed-large + Descriptions & 40.03 & 48.25 & 53.58 & 60.73 & 64.93 \\
\midrule
\highlightrowb Text-embedding-3-small + Tool2Vec & 63.15 & 73.33 & 80.78 & 84.13 & 85.70 \\
Text-embedding-3-small + Descriptions & 41.12 & 54.47 & 57.65 & 65.67 & 68.78 \\
\bottomrule
\end{tabular}
}
}
\end{center}
\label{table:emb-model-ablation}
\end{table*}

\subsection{nDCG Results}
\label{appendix:ndcg_results}
We evaluate nDCG@K, where K is 3, 5, and 7 on the ToolBench dataset and compare our method to the baselines. 
The results are summarized in \autoref{table:toolbench_ndcg}. 
Our method consistently outperform both the retriever introduced in the ToolBench paper, as well as COLT.
\begin{table*}[!t]
\caption{
Comparison of tool retrieval results on the ToolBench dataset. We compared our methods against two baselines: the ToolBench retriever~\cite{qin2024toolllm} and COLT~\cite{qu2024colt}. Evaluation metrics include nDCG@K, where K values are 3, 5, and 7. In the table, N@K stands for nDCG@K. The best-performing method is highlighted in boldface, while the second-best performing method is underlined. We reproduce the ToolBench retriever results based on the original codebase. For the other baseline method, COLT, we report the numbers available in the paper~\cite{qu2024colt}. We observe similar trends when evaluating Recall@K, as shown in \autoref{table:toolbench_results}.
}
\begin{center}
\footnotesize{
\setlength{\tabcolsep}{6pt}{
\begin{tabular}{c|c|c|c|c|c|c|c|c|c}
\toprule
  \multirow{2}{*}{Method}& \multicolumn{3}{c|}{\textbf{ToolBench I1}} & \multicolumn{3}{c|}{\textbf{ToolBench I2}} & \multicolumn{3}{c}{\textbf{ToolBench I3}} \\
 & N@3 & N@5 & N@7 & N@3 & N@5 & N@7 & N@3 & N@5 & N@7 \\
 \cmidrule{1-10}
 ToolBench Retriever & 81.77 & 86.25 & 87.65 & 69.62 & 74.57 & 78.00 & 57.62 & 61.77 & 66.69 \\
 COLT & - & - & - & 78.57 & 82.54 & - & 81.21 & 84.18 & - \\
 \cmidrule{1-10}
 \highlightrow\OURS & 87.30 & 90.28 & 90.71 & 76.13 & 78.79 & 80.20 & 78.62 & 81.83 & 82.91 \\
 \highlightrow\SIDFORMER & \underline{93.10} & \underline{94.40} & \underline{94.62} & \underline{83.70} & \underline{85.22} & \underline{86.03} & \textbf{86.56} & \textbf{86.93} & \textbf{88.02} \\
 
 \highlightrow\RERANKER + \OURS & 91.11 & 92.30 & 92.93 & 83.00 & 84.98 & 85.93 & 82.81 & 84.37 & \underline{86.47} \\
 \highlightrow\RERANKER + \SIDFORMER  & \textbf{93.37} & \textbf{94.96} & \textbf{95.37} & \textbf{84.10} & \textbf{85.62} & \textbf{86.51} & 
 \underline{84.54} & 
 \underline{84.87} & 
 85.43 \\
\bottomrule
\end{tabular}
}
}
\end{center}
\label{table:toolbench_ndcg}
\end{table*}
\section{Detailed Discussion}
\label{appendix:discussion}
In this section, we conduct a series of analyses to investigate why the \RERANKER combined with \OURS performs better than \OURS across all \OURDATA datasets. 
For simplicity, we will refer to \RERANKER combined with \OURS as simply \RERANKER in this section.
Particularly, we aim to pinpoint the specific tools that both methods struggle with, quantify the mistakes, and asses how query complexity affects tool retrieval performance. 
Our results show that \RERANKER is better able to handle complex queries and maintain consistent performance across a diverse set of tools, while \OURS struggles with simpler queries and makes errors on certain tools more frequently. 

In our initial analysis, we aimed to identify the tools that \RERANKER and \OURS most frequently failed to retrieve. Specifically, the model fails to retrieve a tool when the tool is one of the ground truth tools but isn't retrieved. We then divided the number of these failures by the total occurrences of each tool in the dataset to calculate the percentage failure rate for each tool. For all experiments, we retrieved the top-5 tools, which is the maximum number of tools any data point in \OURDATA needs to retrieve.

\begin{figure}[!t]
    \centering
    \begin{minipage}{0.45\textwidth}
        \centering
        \captionof{table}{
        Comparison of the distribution of percentage failure rates of \RERANKER + \OURS and \OURS across all tools in \OURDATA. We calculate the percentage failure rate for a tool as the number of times the method fails to retrieve the tool divided by the number of times it was used in the entire dataset. 
        }
        \scriptsize{
        \setlength{\tabcolsep}{6pt}{
        \begin{tabular}{l|c|c|c|c|c|c}
        \toprule
        \multirow{2}{*}{Data} & \multicolumn{2}{c|}{\textbf{ToolRefiner + Tool2Vec}} & \multicolumn{2}{c|}{\textbf{Tool2Vec}} \\
        \cmidrule{2-5}
         & Mean & Std. & Mean & Std. \\
        \midrule
        \NUMPY & 3.55 & 6.54 & 9.04 & 37.88 \\
        \PANDAS & 2.97 & 6.24 & 12.06 & 26.16 \\
        \AWS & 1.42 & 1.61 & 7.01 & 15.13 \\
        \bottomrule
        \end{tabular}
        }
        }
        \label{table:tool_dist}
    \end{minipage}\hfill
    \begin{minipage}{0.45\textwidth}
        \centering
        \includegraphics[width=\linewidth]{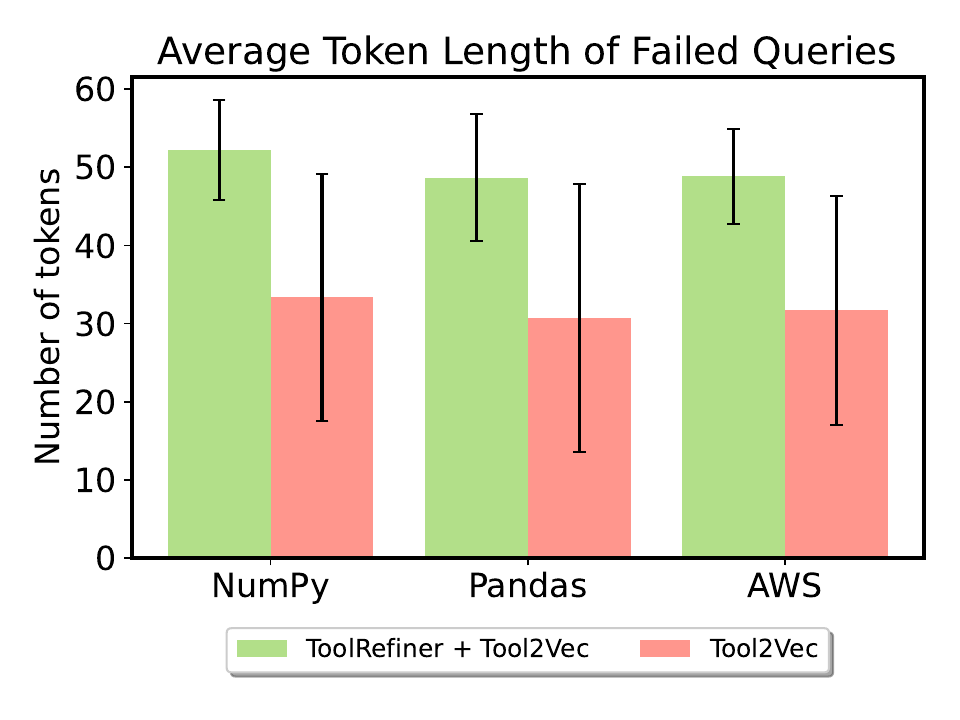}
        \captionsetup{font=small}
        \caption{
        Illustration of average token length of failed queries for \RERANKER + \OURS combined with \OURS and \OURS on \OURDATA is analyzed. 
        We visualize the mean as a bar plot and the standard deviation as an error bar within each bar. 
        We observe that \OURS struggles with shorter and simpler queries, while \RERANKER + \OURS tends to make mistakes on longer and more complex queries.
        }
        \label{fig:query_length}
    \end{minipage}
\end{figure}

In \autoref{table:tool_dist}, we show the distribution statistics of the percentage failure rates of all tools in \OURDATA subsets. We observe that \RERANKER demonstrates a more uniform failure rate distribution, with a relatively low mean and standard deviation. It rarely makes more than five errors per tool and averages 2.07 mistakes across all datasets. 
This suggests that \RERANKER has a robust understanding of a broad range of tools, managing to maintain relatively low failure rates consistently.

On the other hand, \OURS exhibits significant variability in its performance. Certain tools are prone to high failure rates, with some reaching up to 50 mistakes, while others have no errors at all. From \autoref{table:tool_dist}, we observe that \OURS's failure rate distribution is highly skewed, meaning that there are some tools that are responsible for a majority of \OURS's errors.
Furthermore, on average, \OURS makes 7.86 mistakes per tool, indicating a less consistent performance across the board. 
This variability might be due to \OURS's handling of tool embeddings, where it fails to adequately differentiate between tools with similar functionalities. 
In contrast, \RERANKER effectively separates embeddings of tools with overlapping or similar use cases, which can be closely clustered in the \OURS space. 
By diversifying these embeddings, \RERANKER reduces the likelihood of confusion and errors, particularly in complex query scenarios.

In our further analysis in \autoref{fig:query_length}, we focus on the length of the queries where the methods fail and discover that \RERANKER generally makes errors on longer queries, averaging nearly 20 tokens more than those where \OURS failed. 
This finding implies that while \RERANKER is equipped to handle more complex and lengthy queries, \OURS tends to struggle with simpler, shorter queries. 
The ability of \RERANKER to process longer and potentially more complex queries underscores its enhanced capability to manage intricate or verbose user requests effectively.

The disparities in percentage failure rate distribution and the correlation with query length suggest that \RERANKER's superior performance can primarily be attributed to its refined handling of challenging queries and its robustness across a diverse set of tools.

\section{Details on Dataset Generation}
\label{appendix:dataset_generation_details}
\subsection{Prompt for Data Generation}
\label{appendix:prompt_for_data_generation}
\subsubsection{Query Generation}
The following prompt is used to generate user query for \OURDATA.

\begin{tcolorbox}[colback=blue!5!white, colframe=blue!5!white, fonttitle=\bfseries, breakable]
\begin{verbatim}
You are an expert in utilizing a library of functions and generating 
diverse scenarios where a set of selected functions are applied to 
solve real-world problems. You will be provided with a set of 
functions and their descriptions, and will be tasked with selecting 
a subset of these functions to craft detailed scenarios. You will 
generate clear and detailed user instructions, list the names of 
the relevant functions, and explain how these functions can be 
applied to complete the task. These tasks should demonstrate a wide 
range of functionalities and real-life applications to ensure variety 
and utility.

Guidelines:

- The instructions must be clear and comprehensive, allowing users 
  to understand how to apply the functions without ambiguity. However, 
  the instructions shouldn't be robotic and shouldn't sound like 
  'step-by-step' instructions. For example, instead of writing 
  ``Calculate the non-negative square root of an array element-wise, 
  then round the resulting array to the nearest even value, and return 
  the indices that would sort the array along a specified axis.'' which 
  breaks down each step mechanically, you MUST instead write a more 
  natural and fluid instruction like ``Sort the array along a specified 
  axis after calculating the non-negative square root of each element 
  and rounding the result to the nearest even value.''

- You MUST select and sequence the functions in a way that demonstrates 
  their interdependency. Ideally, a function's output should be the input 
  to another function (or multiple functions), creating a chain of 
  operations that solve the task at hand. In other words, the functions 
  you select must not be selected randomly but instead be used to solve 
  coherent multi-step problems. 

- The explanations should logically connect the functions to the tasks, 
  demonstrating the workflow clearly. 

- Your response should be returned as a single JSON object, representing 
  a unique user instruction. Diversity in function use and application 
  context is crucial; avoid repetition of similar tasks or functional 
  applications to ensure a broad coverage of the capabilities of the 
  functions. 

Here is an example output of a list of JSON objects representing very 
distinct and detailed tasks:

``{examples_str}''

You MUST only return a single JSON object - do not add any extra text 
before and after the json object. The instructions that you generate 
MUST be very diverse and distinct from each other and MUST be as different 
as possible from the examples above. ``{library_specific_instructions}''
\end{verbatim}
\end{tcolorbox}

\subsubsection{Query Polish}
The following prompt is used to polish user query for \OURDATA.
\begin{tcolorbox}[colback=blue!5!white, colframe=blue!5!white, fonttitle=\bfseries, breakable]
\begin{verbatim}
You are an expert at refining user instructions to make them more coherent 
and less robotic. You will be given a user instruction and will be tasked 
to refine the user instruction if it:

- Sounds too robotic or step-by-step like saying 'Do this, do that, and 
then do this'. In other words, the instructions shouldn't break down each 
step  mechanically but be more fluid. For example, instead of writing 
"Analyze the  lyrics of the song 'XYZ', generate a playlist based on 
the emotions and themes found, and create a Spotify playlist with 
the recommended songs." you would write "Create a Spotify playlist based 
on the emotions and themes found in the lyrics of the song 'XYZ'.

- Has conditional statements like 'if this, then do that' or 'when this 
happens, do that'. It should be more direct and non-conditional.

If none of the above applies to an instruction, you should mark it as good, 
and provide a reasoning for why it is good. Here example outputs of a JSON 
object representing a refined user instruction:

``{in_context_examples}''
\end{verbatim}
\end{tcolorbox}

\subsection{Tool Selection Criteria}
For tool collection, we crawl each library's official API reference and retrieve detailed information about function descriptions, arguments, and example code snippets.
For NumPy, we exclude the numpy.ctypeslib, numpy.dtypes, numpy.emath, numpy.rec, and numpy.version modules since they don't provide rich functions or are outdated. 
For Pandas, we only use the public sub-packages and exclude the pandas.core, pandas.compat, and pandas.util modules. 
For Boto3, we include functions for five popular AWS services: EC2, RDS, IAM, S3, and SNS. 

\subsection{Parameters and LLMs to Generate Dataset}
For dataset generation, we used \( T=10 \) and \( M=2-5 \), which means that at each iteration, we sample 10 tools from the tool pool and let the language model choose 2-5 tools to generate instructions. 
For both the Instruction Generation and Instruction Polish stages, we use Llama-3-70B-Instruct ~\cite{llama3modelcard}. 

\subsection{Dataset Statistics}
We collect 511 NumPy tools, 1655 Pandas tools, and 1002 Boto3 tools and curate about 20,000 NumPy queries, 70,000 Pandas queries, and 73,000 AWS queries. There are 19530 tool combinations in our NumPy dataset, 69550 in Pandas dataset, and 70816 in AWS dataset. This means that almost all of our data represents distinct usage scenarios and queries. 

\begin{figure*}[!t]
    \centering
    \captionsetup{font=small}
    \includegraphics[width=\linewidth]{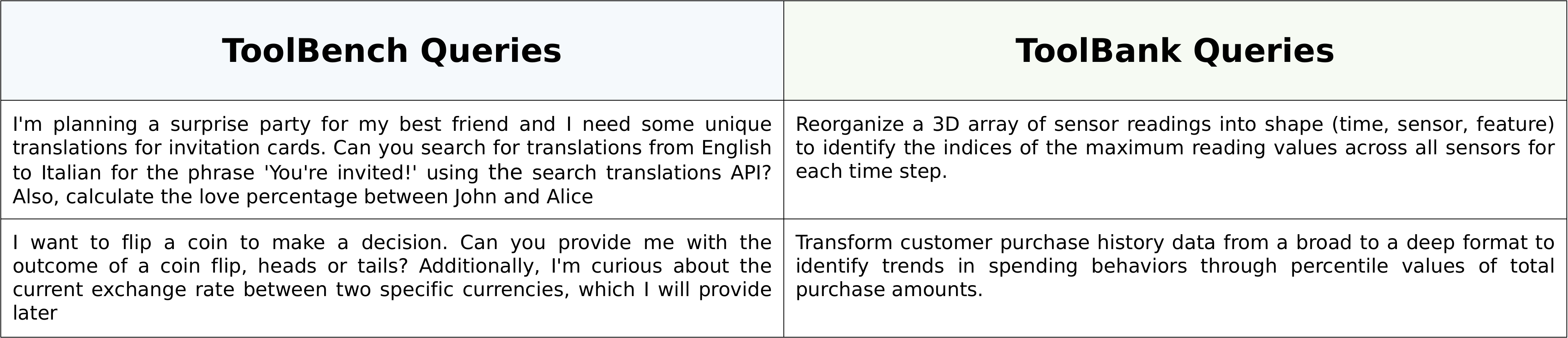}
    \caption{
    Qualitative analysis comparing the queries in ToolBench and \OURDATA. We randomly sample 2 examples from each dataset.
    Queries in ToolBench often follow an artificial pattern like "Do this, do this, and do this," resulting from random sampling of multiple tools from RapidAPI Hub. In contrast, \OURDATA queries are more natural, resembling real human queries to LLMs, with coherent and related tools better aligned to user needs.
    }
    \label{fig:qualitative_analysis}
\end{figure*}

\subsection{Qualitative Analysis}
\subsubsection{Comparing \OURDATA and ToolBench}
Continuing the discussion on \autoref{sec:tool_retrieval_dataset}, we present a qualitative comparison between \OURDATA and ToolBench. 
We observe that the format of queries in ToolBench often follows the pattern ``Do this, do this, and do this," which results from randomly sampling multiple tools from RapidAPI Hub. 
This format is somewhat artificial compared to how real human users give queries to LLMs for certain tasks. 
Additionally, some queries directly or indirectly mention the required APIs, simplifying tool retrieval. 
In contrast, the queries sampled from \OURDATA are more natural, closely resembling how real human users are likely to ask LLMs to perform tasks.
Furthermore, the tools required for each user query task in \OURDATA are more coherent and related to each other, ensuring better alignment with user needs.
For further detail, please refer to \autoref{appendix:dataset_generation_details}.

\subsubsection{Comparing Polished and Unpolished Queries}
\begin{figure*}[h]
    \centering
    \captionsetup{font=small}
    \includegraphics[width=\linewidth]{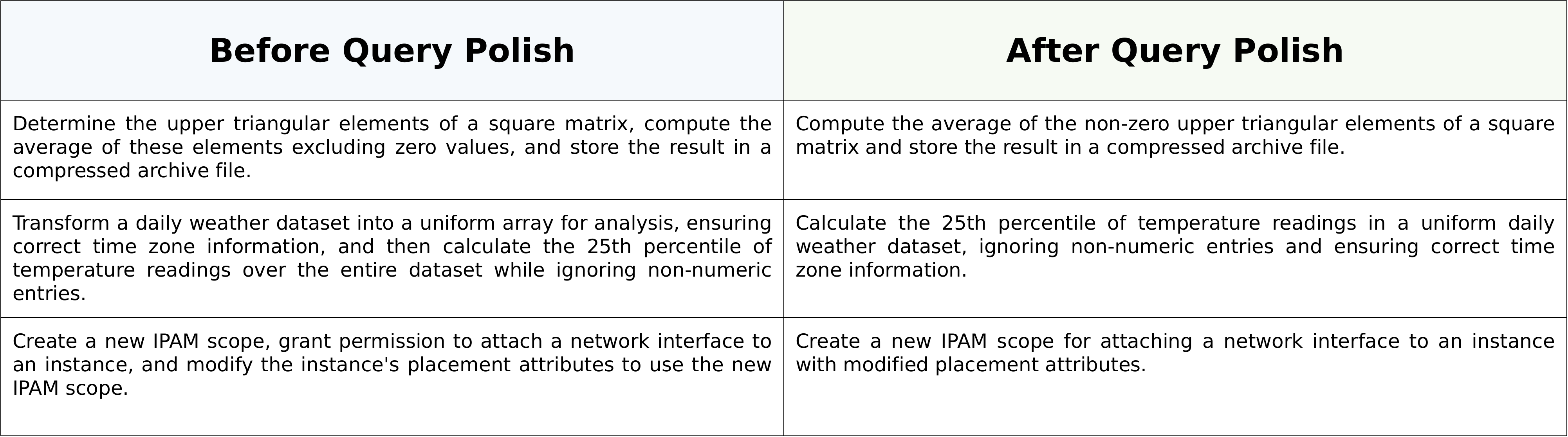}
    \caption{
Qualitative analysis comparing the queries before and after the Query Polish stage. We randomly sample an example from each of \NUMPY, \PANDAS, and \AWS datasets.}
    \label{fig:unpolished_instructions}
\end{figure*}

We provide a qualitative analysis of \OURDATA and investigate the effect of the Query Polish step. We randomly sample three queries from \OURDATA and present them before and after the polishing step in \autoref{fig:unpolished_instructions}.
The left column in \autoref{fig:unpolished_instructions} shows the queries before applying the Query Polish step, while the right column shows the queries after polishing. 
Before applying Query Polish, the queries exhibit a rigid and instructional style, similar to the ToolBench examples in \autoref{fig:qualitative_analysis}. 
However, after applying Query Polish, the queries become more natural and user-friendly, better reflecting how real human users would interact with LLMs.

\subsubsection{Qualitative Analysis on \OURDATA}
\label{appendix:pandas_worse}
In this section, we compare \NUMPY, \PANDAS, and \AWS, which are subsets of \OURDATA, and provide insights into why the tool retrieval performance on \PANDAS is worse than on the other subsets. This performance degradation can largely be attributed to the similarity between operations in \PANDAS. Specifically, \PANDAS contains various data types like time series, periods, intervals, and indexes. For these data types, there are some common set of operations that apply to all of them such as $\texttt{.equals}$, $\texttt{.argmin}$, or $\texttt{.all}$. This results in instructions that are very close to each other in meaning but use different data types. Hence, the model gets confused about which data type to operate on. For example, the first part of ``Load a delimited data file with specific columns and data types, counting the total number of entries, for the next fiscal quarter starting from the first business day of the year based on a given timestamp.'' query requires a call to $\texttt{pandas.read\_csv}$ tool which returns a $\texttt{pandas.DataFrame}$ object and a subsequent call to $\texttt{pandas.DataFrame.size}$ tool to count the total number of entries. However, \RERANKER + \OURS model makes the mistake of calling $\texttt{pandas.Series.size}$ after calling $\texttt{pandas.read\_csv}$. This doesn't become an issue with \NUMPY and \AWS since \NUMPY tools operate almost entirely on the array type and \AWS contains only the 5 most popular AWS services, creating a clear distinction between tools.


\end{document}